\documentclass{article}

\usepackage{arxiv}

\usepackage[utf8]{inputenc} 
\usepackage[T1]{fontenc}    
\usepackage{hyperref}       
\usepackage{url}            
\usepackage{booktabs}       
\usepackage{amsfonts}       
\usepackage{nicefrac}       
\usepackage{microtype}      
\usepackage{lipsum}		
\usepackage{graphicx}
\usepackage{natbib}
\usepackage{doi}

\usepackage{algpseudocode}
\usepackage{algorithm}
\usepackage{amsmath}
\algnewcommand\algorithmicassume{\textbf{Assume:}}
\algnewcommand\Assume{\item[\algorithmicassume]}

\usepackage{tabularx}

\newenvironment{conditions*}
  {\par\vspace{\abovedisplayskip}\noindent
   \tabularx{\columnwidth}{>{$}l<{$} @{${}={}$} >{\raggedright\arraybackslash}X}}
  {\endtabularx\par\vspace{\belowdisplayskip}}

\usepackage[scientific-notation=true, round-precision=3,round-mode=figures]{siunitx}
\usepackage{graphicx}

\usepackage{svg}

\usepackage{placeins}

\title{Streaming Bayesian Modeling for predicting Fat-Tailed Customer Lifetime Value.}

\author{
\hspace{1mm}Alexey V. Calabourdin\\
	Engineering School of Information Technologies, \\
        Telecommunications and Control Systems. \\
	Ural Federal University, Yekaterinburg. \\
	\texttt{a.calabourdin@gmail.com} \\
 \And
\hspace{1mm}Konstantin A. Aksenov\\
	Engineering School of Information Technologies, \\
        Telecommunications and Control Systems. \\
	Ural Federal University, Yekaterinburg. \\
	\texttt{k.a.aksenov@urfu.ru} \\
}

\renewcommand{\shorttitle}{\textit{arXiv} Template}

\hypersetup{
pdftitle={Streaming Bayesian Modeling for predicting Fat-Tailed Customer Lifetime Value.},
pdfsubject={cs.LG},
pdfauthor={Alexey Calabourdin},
pdfkeywords={stream learning, online learning, data stream, concept drift, hierarchical bayesian modeling, mcmc, monte-carlo, lifetime value, fat tails},
}

\begin{document}
\maketitle

\begin{abstract}
        We develop an online learning MCMC approach applicable for hierarchical bayesian models and GLMS. We also develop a fat-tailed LTV model that generalizes over several kinds of fat and thin tails. We demonstrate both developments on commercial LTV data from a large mobile app.
\end{abstract}


\section{Introduction}
\label{sec:Introduction}
\begin{figure}[h]
   \centering
   \begin{tabular}{cc}
       \includegraphics[width=0.45\textwidth]{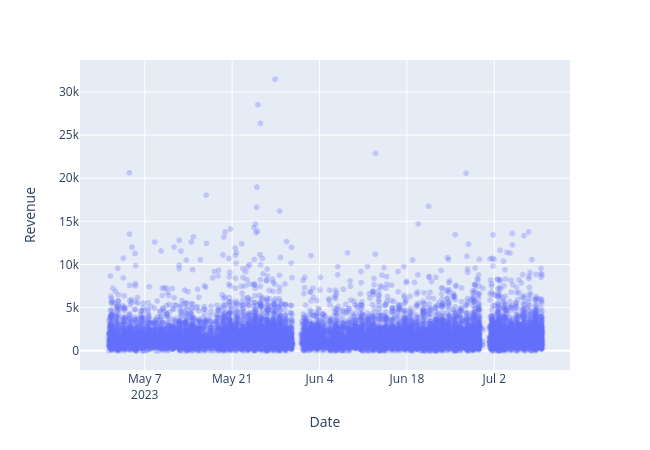} &
       \includegraphics[width=0.45\textwidth]{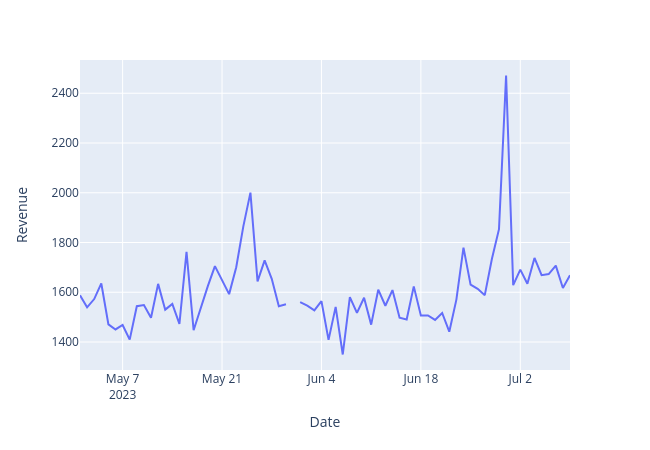}
   \end{tabular}
   \caption{Mysterious revenue dataset. Individual datapoints (left), daily mean (right).}
   \label{fig:mysterious-revenue-dataset}
\end{figure}

Consider machine learning modeling for example dataset in Figure \ref{fig:mysterious-revenue-dataset}.

The distribution exhibits occasional highly skewed variance, it doesn’t appear stationary
\footnote{The reader can assume, perhaps, monthly seasonality/cycles, but for generality let’s not assume.}. 
Consider modeling for high volumes of such data, larger than memory volumes. Consider accounting for high-dimensional feature sets, e.g. multiple sparse high cardinality categorical features ($10^4+$ cardinality). Consider higher variance, more drastic data drifts - and consider impactful financial consequences from over-/underestimating them. Consider a necessity to update such a model frequently, e.g. daily.

Figure \ref{fig:mysterious-revenue-dataset} shows a sample of user Lifetime Value (LTV) data expressed in revenue. 18 000 freshly acquired users of a mobile app, each point representing a day 7 cohort LTV for each unique user. Similar datasets can be found in energy cost prediction, in stock exchange rate prediction, in retail sales prediction.

For such a problem, a bayesian modeling approach can be helpful. Bayesian modeling allows to estimate probability distributions of predicted variables, enabling decision-making under uncertainty. Compared to methods producing only point forecasts this gives an advantage in modeling fat-tailed variables \citep{taleb2022single} like in our problem. Arguably the most expressive bayesian models are based on Markov Chain Monte Carlo (MCMC).

But bayesian modeling, especially MCMC, is challenging to scale to wide data (lots of features) and tall data (lots of observations), compared to frequentist methods. It’s also more challenging to apply in online learning settings and adapt to concept drifts. In fact, we couldn’t even find any examples of online MCMC modeling in popular PPL documentations, such as NumPyro\footnote{https://github.com/pyro-ppl/numpyro}, Turing.jl\footnote{https://github.com/TuringLang/Turing.jl}, PyMC\footnote{https://github.com/pymc-devs/pymc}. And accounting for concept drifts can be non-straightforward, as we discuss in Section \ref{sec:Converting Bayesian MCMC Models to Online Models}.

In this work we develop a method to make MCMC bayesian modeling applicable:
\begin{itemize}
\item At scale (for tall/wide data)
\item In online learning settings
\item Under concept drifts
\end{itemize}

We also develop a fat-tailed Lifetime Value (LTV) model leveraging this approach.

Specifically, we combine several methods:
\begin{itemize}
\item Online Machine Learning
\item Bayesian Modeling
\item Estimation with Student-$t$ distributions
\end{itemize}

Notable contributions of our work, inspired by practical challenges:
\begin{itemize}
\item We develop a reusable workflow for converting batch models to mini-batch models for MCMC-based bayesian models. This workflow is applicable for popular methods such as Generalized Linear Models (GLM) and Hierarchical Bayesian Models. This workflow allows to retain model interpretability, causal and hierarchical relationships in the aforementioned methods. To make this workflow work, we develop:
    \begin{itemize}
    \item An extension to the traditional ordinal category encoding algorithm to make it online and “infinitely” extensible. This enables straightforward online preprocessing for index encoding, the go-to categorical encoding algorithm in bayesian modeling.
    \item An algorithm to convert HMC/NUTS sampler to online HMC/NUTS sampler in a way that is robust to concept drifts. This allows us to leverage tested popular probabilistic programming frameworks like NumPyro for online bayesian modeling under concept drifts.
    \end{itemize}
\item We develop a model for user LTV using our workflow. The resulting model generalizes for LTV returns from different kinds of probability distribution tails, including thin tails of Gaussian distribution and extremely fat tails of Cauchy distribution. The model is adaptable to concept drifts in data. The model can be updated without retraining on historic data. The model can be trained on larger than memory volumes of data.
\end{itemize}

To our knowledge this is the first combination of online learning, MCMC-based bayesian modeling and a fat-tailed variable estimation for LTV prediction in Adtech having the aforementioned features.

Overview and comparison of aspects of our contribution with existing developments are presented in \nameref{sec:Related work} section.

\section{Converting Bayesian MCMC Models to Online Models}
\label{sec:Converting Bayesian MCMC Models to Online Models}

Assume this intuitive definition for online machine learning task:

\emph{Online machine learning} - learning models incrementally from data in a sequential manner. 

Assume extra challenges:
\begin{itemize}
\item \emph{Stream Learning} \citep{gama2013evaluating, domingos2001catching}. We imply 1-pass learning, i.e. learning from each data point only once, without retaining historic data and doing multiple epochs of learning steps.
\item \emph{Adaptive Learning} \citep{hoi2021online}. We imply adaptation to concept drifts in data. I.e. we assume data is not stationary.
\end{itemize}

These challenges are to demonstrate a more general applicability of our approach.

Consider a simple frequentist machine learning model, a linear regression with SGD-based optimizer, with optional L1/L2 regularization. Such model is a reasonable baseline for the problem from \nameref{sec:Introduction}. A popular library for machine learning, Scikit-Learn\footnote{https://github.com/scikit-learn/scikit-learn}, provides a reference implementation with good engineering quality. 

Now consider an online version of such a model. A popular library for online machine learning, River \footnote{https://github.com/online-ml/river}, provides a reference implementation with good engineering quality.

To convert from batch to online learning, the online version implements:

\begin{itemize}
\item Online preprocessing
    \begin{itemize}
    \item For continuous features
    \item For categorical features
    \end{itemize}
\item Incremental updates for the model and preprocessing states
\end{itemize}

Now consider a simple bayesian MCMC version of linear regression (example can be found in \citep{mcelreath2018statistical}). 

To convert from batch to online learning, we just as well need:

\begin{itemize}
\item Online preprocessing
    \begin{itemize}
    \item For continuous features
    \item For categorical features
    \end{itemize}
\item Incremental updates for the model and preprocessing states
\end{itemize}

Online preprocessing for continuous features can be reused from frequentist workflow (standard scaling, robust scaling can be seamlessly reapplied). 

Online preprocessing for categorical features we need to address. Ideally we would like to accomodate index variable encoding, a flexible way of encoding categoricals in bayesian models instead of one-hot encoding \citep{mcelreath2018statistical}.

Incremental updates for preprocessing states can be reused from frequentist workflow.

Incremental updates for the model we need to address, because MCMC models are based on different optimizer with different behavior. Ideally we’d like to accommodate some very sample-efficient and time-tested sampler which can provide very high quality estimates, for example HMC\citep{neal2011mcmc} or HMC/NUTS\citep{hoffman2014no} sampler.

Bayesian linear regression model can be extended to GLM and Bayesian Hierarchical Model. Thus our batch to online conversion workflow too can be extended to GLM and Bayesian Hierarchical Model cases. In turn, extending to Hierarchical Bayesian Models \citep{mcelreath2018statistical} allows us to regularize and deal with sparsity via hierarchical priors, as well as encode causal relationships. 

\subsection{Streaming Categorical Encoding}
\label{subsec:Streaming Categorical Encoding}

\paragraph{Definition}

Intuition for our proposed streaming categorical encoding: 

\emph{Encode categorical features as an integer array, where category codes are assigned based on chronological order of the first occurence of category values in data.}

In a more formal way, the classic proper use of ordinal encoding algorithm can be defined as follows:

\begin{algorithm}
\caption{Ordinal Encoding Algorithm (single feature example)}
\begin{algorithmic}[1]
\Assume Ordering of category values based on value semantics 
\Require 
\State Order-preserving mapping $M : \textbf{v} \mapsto \textbf{c}$ for a list of known unique category values $\textbf{v} = [v_1, v_2, ... v_m]$ to a list of unique integer codes $\textbf{c} = [1, 2, ... m]$, where $\textbf{v}$ is sorted in ascending order according to semantics of its values.
\State Batch of data $\textbf{X}$  with $n$ rows, containing category values.
\Ensure Batch of transformed data $\textbf{X}_{tr}$ with $n$ rows, containing encoded category values.
\For{$i \gets 1$ to $n$}
    \If{$x_i \in M$}
        \State $x_{itr} \gets M(v)$
    \Else
        \State $x_{itr} \gets c_{unknown}$
    \EndIf
\EndFor
\State \Return $\textbf{X}_{tr}$
\end{algorithmic}
\end{algorithm}

The proposed streaming ordinal encoding algorithm:

\begin{algorithm}
\caption{Streaming Ordinal Encoding Algorithm (single feature example)}
\begin{algorithmic}[1]
\Assume Ordering of category values based on the order of occurrence in the data stream. 
\Require 
\State Stream of data $\textbf{X}$  with $n$ rows, containing category values
\Ensure Stream of transformed data $\textbf{X}_{tr}$ with $n$ rows, containing encoded category values
\State Initialize empty mapping $M : \textbf{v} \mapsto \textbf{c}$ with $c_{unknown} \gets 0$ \Comment{reserving first codes for special purposes}
\For{$i \gets 1$ to $n$}
    \If{$x_i \in M$}
        \State $x_{itr} \gets M(v)$
    \Else
        \State $x_{itr} \gets c_{unknown}$
        \State $M(v_i) \gets M.length + 1$
    \EndIf
\EndFor
\State \Return $\textbf{X}_{tr}$
\end{algorithmic}
\end{algorithm}

\paragraph{Commentary}
We call our algorithm ordinal encoding. But, strictly speaking, the order here represents not the intrinsic relation between category values (as intended by classic ordinal encoding algorithm), rather, it represents the observer’s subjective perception of category values. So one might also call it “chronological label encoding”.

\paragraph{Application details}
Integer codes from this encoding serve as indices in index variable encoding for respective category values. This is convenient because the highest code represents the cardinality seen so far plus, optionally, unknown/empty values, so we don’t get out of bounds in index variable encoding. Provided we provision enough encoding space in index variable encoding in advance, before training a model, to accommodate the full cardinality. Practically this can scale to at least tens of thousands of category values with no significant overhead. The limitation is that due to specifics of modern PPLs provisioning is done using fixed data structures (arrays, tensors), so once the current index variable encoding table is full, the model needs to be refreshed / encoding space increased.

Comparing to classic label encoding (for implementation see Scikit-Learn) we don’t have random assignment of values, which makes it straightforward to update to new data with new category values and combine with bayesian index variable encoding. 

Comparing to classic ordinal encoding (for implementation see Scikit-Learn) we are explicit about having chronological codes and provide incremental update functionality. Nevertheless, in batch setting this encoder can be parameterized with explicit chronological codes to reproduce the same encoding scheme as ours. This can be handy to leverage Scikit-Learn’s rich ecosystem and tool integrations for engineering purposes, especially in production environments.

We implement the algorithm as an extension of River library. A polished implementation is available in the current River release courtesy of Max Halford (one of the River core devs).

\subsection{Online HMC/NUTS}
\label{subsec:Online HMC/NUTS}

\paragraph{Definition}
Intuition for our proposed online HMC/NUTS algorithm:

\emph{Do sampling for each mini-batch, reuse the state from previous sampling and do extra warmup before fitting to mini-batch.}

The classic modeling workflow with HMC/NUTS sampler (as implemented in PPLs like NumPyro):
\begin{algorithm}
\caption{Classic modeling with HMC/NUTS sampler}
\begin{algorithmic}[1]
\Assume Modeling regression of $\textbf{X}$ on $\textbf{y}$
\Require
\State Data $\textbf{X}$ with $n$ rows.
\State Log-probability density function ${\hat{y}(X)}$. \Comment{referred to as "the model" in practice}
\State Sampler parameters, including number of samples $s$ and number of warmup steps $w$.
\Ensure $n \times s$ posterior predictive samples $\textbf{\^{Y}}$.
\State Initialize the sampler
\State Perform $w$ warmup steps
\For{$i \gets 1$ to $s$}
    \State "Fit" the HMC/NUTS sampler and infer posterior samples with ${\hat{y}(X)}$. This step requires scanning over $n$ rows.
\EndFor
\State Infer $n \times s$ posterior predictive samples $\textbf{\^{Y}}$ \Comment{basically our predictions, in the form of numerical densities}
\State Score the predictions for $\textbf{\^{Y}}$
\State \Return $n \times s$ posterior predictive samples $\textbf{\^{Y}}$.
\end{algorithmic}
\end{algorithm}

The proposed mini-batch modeling workflow with HMC/NUTS sampler:
\begin{algorithm}
\caption{Proposed mini-batch modeling workflow with HMC/NUTS sampler}
\begin{algorithmic}[1]
\Assume Modeling regression of $\textbf{X}$ on $\textbf{y}$
\Require
\State Data $\textbf{X}$ with $n$ rows split by $k$ mini-batches $\textbf{x}$.
\State Log-probability density function ${\hat{y}(X)}$. \Comment{referred to as "the model" in practice}
\State Sampler parameters, including number of samples $s$ and number of warmup steps $w$.
\State Number of extra warmup steps $w_{extra}$
\Ensure $n \times s$ posterior predictive samples split by $k$ mini-batches: $\textbf{\^{Y}}$ = [$\textbf{\^{y}}_1$, $\textbf{\^{y}}_2$, ... $\textbf{\^{y}}_k$].
\State Initialize the sampler
\State Perform $w$ warmup steps on the first batch $\textbf{x}_{1}$
\State Infer $\frac{n}{k} \times s$ posterior predictive samples for the first batch $\textbf{x}_{1}$
\State Extract the sampler state $state_{last} \gets state_{1}$
\For{$j \gets 2$ to $k$}
    \State Infer posterior samples with ${\hat{y}(X)}$. Each sample requires scanning over $\frac{n}{k}$ rows.
    \State Infer and collect $\frac{n}{k} \times s$ posterior predictive samples $\textbf{\^{y}}$ \Comment{basically our predictions for mini-batch $j$, in the form of numerical densities}
    \State Score the predictions
    \State Perform $w_{extra}$ warmup steps on the batch $\textbf{x}_{j}$, sampler initialized with $state_{init} \gets state_{last}$
    \State Extract the sampler state $state_{last} \gets state_{j}$
    \State "Fit" the HMC/NUTS sampler on the batch $\textbf{x}_{j}$, sampler initialized with $state_{init} \gets state_{last}$
    \State Extract the sampler state $state_{last} \gets state_{j'}$
\EndFor
\State \Return $n \times s$ posterior predictive samples split by $k$ mini-batches: $\textbf{\^{Y}}$ = [$\textbf{\^{y}}_1$, $\textbf{\^{y}}_2$, ... $\textbf{\^{y}}_k$].
\end{algorithmic}
\end{algorithm}

\paragraph{Commentary}
Extra warmup enables usage under concept drifts. Otherwise the model effectively stops learning if the concept drifts are drastic (e.g. x20 shift in target distribution mean). This can supposedly be seen as effectively a re-adaptation of HMC mass matrix to new potential energy due to changes in posterior geometry. In terms of a frictionless puck analogy\citep{neal2011mcmc}, as if the puck was put into a slightly different surface at different height and hamiltonian dynamics state estimates need to be updated accordingly.

\paragraph{Application details}
In practice we’ve found that taking multiple samples (e.g. 1000+ samples) over the same mini-batch provides good modeling results. This assumes multiple HMC proposals, multiple trajectories and multiple leapfrog steps per minibatch. However for some applications taking fewer samples might be more desirable to avoid overfitting to particularly noisy/stochastic data. Reducing it to 1 sample over the minibatch is not recommended unless the reader intends to do multiple passess (epochs) over the whole datasets (effectively abandoning the Stream Learning and Adaptive Learning challenges). This would make the proposed approach similar to mini-batching approach proposed in classic work \citep{neal2011mcmc} or popular in Bayesian Neural Network community Langevin dynamics / Stochastic Gradient MCMC family methods like SGLD, SGHMC and others \citep{welling2011bayesian, chen2014stochastic, vadera2020ursabench}. If the reader doesn’t consider addressing Stream Learning and Adaptive Learning challenges they are recommended to explore the body of work on these methods. A good quality reusable implementation of SGLD is available in BlackJax library\footnote{https://github.com/blackjax-devs/blackjax}.

The amount of extra warmup steps sufficient varies. In our experiments we’ve found the number of samples x3 to be a working rule of thumb. Further implications of this heuristic remain to be studied.

For the mini-batch size we’ve found samplers from NumPyro and BlackJax to be quite robust to mini-batch size increase in terms of memory footprint. But the reader also wouldn't want to set mini-batch size too large otherwise they loose on concept drift adaptation capabilities. For datasets with number of observations in the order of magnitude $10^{4} - 10^{6}$ we’ve found mini-batch size in the order of $10^3$ to be a reasonable default. This more or less aligns with our experience in frequentist SGD-based online learning algorithms.

\section{Modeling LTV with Fat Tails}
\label{sec:Modeling LTV with Fat Tails}
LTV modeling is a problem of modeling lifetime value of customers to inform business decision making. Customer LTV can be modeled as a random variable from some probability distribution. I

The specific details of the LTV problem we consider can be found at the end of this section, for the sake of explaining the model they are not essential.

Now assume a simple definition of fat tailed distribution inspired by \citep{taleb2018much}:

\emph{A fat tailed distribution is a distribution with tails fatter than tails of Gaussian distribution.}

This implies extreme events are more likely than under Gaussian distribution. Thus Gaussian distribution sometimes said to have thin tails.

As evident in literature, practicioners have noticed the fat-tailed nature of LTV returns - the fact that extraordinary high returns are more likely in practice than under Gaussian distribution assumptions \citep{Wang2019ADP}. By identifying LTV prospects with potential fat-tailed returns a decision maker can increase business revenue by orders of magnitude. 

Now, should we assume every customer from our problem has fat-tailed LTV returns? Not necessarily. Fat-tailed LTV returns could come from:
\begin{itemize}
    \item All of the customers
    \item None of the customers
    \item Some of the customers
\end{itemize}

Certain business domains can be described by Pareto metaphor ("80\% of revenue is generated by 20\% of customers"). We consider the problem from mobile apps business and it’s one of such domains. 

Also not all fat tails are created equal. Some are more fat, some are less fat. See discussion in \nameref{sec:Related work}.

The intuition for our solution is as follows:

\emph{Find a distribution that generalizes over very fat tails and very thin tails. Use it as a distribution for target variable. Let the model show which prospects are likely to exhibit more fat-tailed behavior or more thin-tailed behavior.}

In fact, such a distribution exists. Student-$t$ distribution, which generalizes over Gaussian distribution (thin tails) and Cauchy distribution (fat tails). It can be parameterized by location $\mu$, scale $\sigma$ and degrees of freedom $\nu$. Notably, $\nu  = \infty$ corresponds to Gaussian distribution, $\nu  = 1$ corresponds to Cauchy distribution.

Student-$t$ distribution is often used in statistical modeling to make the regression estimate more robust to outliers. Our application has other implications, because we’re interested in leveraging those outliers (the extreme events), interested in estimating potential for their occurrence from different prospects, interested in quantitative estimation and comparison.

Bayesian modeling allows us to be very flexible with estimating Student-$t$ distributed variables. We can leverage regularizing / informative priors to impose conservative behavior, e.g. assume more thin-tailed or more fat-tailed version of Student-$t$ by the default. In this problem we assume thin-tailed LTV returns by default to not make our estimates too optimistic. In business decision making this allows us to distribute higher marketing budgets only to the prospects we are more confident becoming high-LTV customers.

More formally:

\textbf{Assume} for prospect $i$ LTV estimate is distributed as Student-$t$, parameterized by $\mu$, $\sigma$, $\nu$.\\
\textbf{Then} higher $\nu$ values are associated with higher LTV potential for prospect $i$.

Since fat-tailed distributions are harder to estimate, the user of such approach should also pay attention to uncertainty in Student-$t$ parameter estimates, which in Bayesian modeling are commonly expressed via credibility/compatibility intervals, HDPI \citep{mcelreath2018statistical}.

\paragraph{Details of the LTV problem we consider}

The example problem we consider is from user acquisition in digital marketing domain. In digital marketing efficient LTV modeling can enable value-based pricing in Real-time Bidding (RTB) advertisement auctions, increasing such business metrics as Return on Ad Spend (ROAS) and Average Revenue per User (ARPU). 

We consider modeling cohort LTV for new paying users of mobile applications. Paying users are the users who have paid at least once. Cohort LTV can be defined as LTV realized by a cutoff day after a user has been properly acquired (installed a mobile application). It can be expressed as follows:
\begin{equation}
LTV_D =  \sum_{d=1}^{D} v_d
\end{equation}
where:
\begin{conditions*}
D & Day of cohort (day since the app installation) \\
LTV_D & Cohort LTV realized by D \\
v_{d} & Value from user realized within day d, total \\
\end{conditions*}

Assume user churn rate and costs associated with acquiring and maintaining the user negligible. Then our LTV problem reduces to a problem of modeling cohort revenue, which is potentially a fat-tailed variable.

\section{Experiments}
\label{sec:Experiments}

We demonstrate our approach on a simple minimalistic LTV model example, where we try to predict cohort revenue for each user. We compare Gaussian LTV returns model and Student-$t$ LTV returns model.

We evaluate distribution fitness and typical error metrics for this task like MAE, RMSE.

We use real data of a single cohort revenue from a large mobile app with in-app purchases, sourced from commercial production environment available to us. 

The data consists of 18 000 observations of unique freshly acquired users with features related to user, product and the context of user acquisition. For each user there’s a $D = 7$ cohort revenue which we aim to model. For ease of demonstration we consider a model based only on “AppCategory” feature. “AppCategory” feature in our dataset is a marketplace category of an app (e.g. “Weather”, “Shopping”, “Travel”, “Social”), where the user saw the ads of our mobile app and thus consitutes a context of user acquisition. But the case can be trivially extended for larger feature sets.

\begin{figure}[!htbp]
    \centering
    \includegraphics[width=0.7\linewidth]{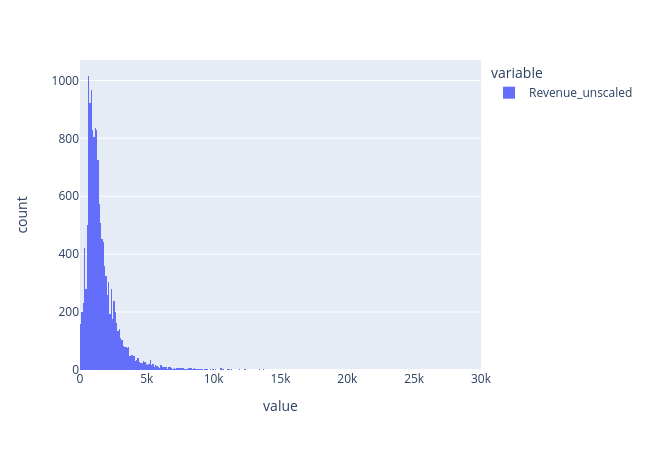}
    \caption{Aggregate revenue distribution.}
    \label{fig:experiments-revenue-data-3}
\end{figure}

For data summary see Table \ref{table:experiments-data-summary-1}, Figure \ref{fig:experiments-revenue-data-3}.

\FloatBarrier

       

\setlength{\tabcolsep}{0pt} 
\renewcommand{\arraystretch}{0.7} 
\begin{table}[!htbp]
\resizebox{0.8\textwidth}{!}{\begin{minipage}{\textwidth}
\begin{tabular}{@{}lSSSSSSSS[table-align-exponent = false]@{}}
\toprule
 & {count} & {mean} & {std} & {min} & {25\%} & {50\%} & {75\%} & {max} \\
{AppCategory} & & & & & & & & \\
\midrule
3 & 5196.000000 & 1600.483854 & 1500.877856 & 1.046166 & 784.212198 & 1211.204315 & 1938.463517 & 28518.255754 \\
4 & 4120.000000 & 1652.068157 & 1484.380054 & 1.039196 & 813.122967 & 1250.282137 & 1963.486650 & 20582.926137 \\
2 & 2823.000000 & 1567.840334 & 1324.434087 & 1.075194 & 785.585315 & 1248.358901 & 1903.852383 & 14689.015975 \\
6 & 1384.000000 & 1672.531231 & 1471.123270 & 1.061407 & 842.555614 & 1294.176869 & 2059.606395 & 26351.946702 \\
9 & 755.000000 & 1730.294675 & 1799.845614 & 1.039527 & 828.567930 & 1358.638947 & 2098.428069 & 31473.388661 \\
5 & 184.000000 & 1384.303192 & 1069.775517 & 1.079972 & 722.220565 & 1067.457255 & 1769.831847 & 7401.731970 \\
8 & 152.000000 & 2043.975885 & 1798.630233 & 1.075302 & 887.175830 & 1475.632627 & 2589.622884 & 12448.075992 \\
11 & 112.000000 & 1424.791268 & 1004.722681 & 1.079568 & 804.811161 & 1216.440090 & 1605.140843 & 6005.587500 \\
7 & 103.000000 & 1786.737631 & 1750.497049 & 123.762892 & 839.127490 & 1229.049297 & 2022.334716 & 12361.979751 \\
15 & 79.000000 & 1359.276291 & 1070.758597 & 1.069358 & 762.568710 & 1023.659659 & 1829.478383 & 6717.962989 \\
18 & 27.000000 & 1645.222922 & 1006.773078 & 197.429596 & 878.421633 & 1422.143643 & 2217.469301 & 3908.779333 \\
16 & 18.000000 & 1597.291228 & 900.304658 & 124.320329 & 1023.840054 & 1414.784710 & 2093.651474 & 3462.971429 \\
13 & 15.000000 & 1763.196020 & 1304.645580 & 608.038042 & 960.638229 & 1456.741572 & 2100.010136 & 5809.091700 \\
12 & 11.000000 & 1716.176543 & 1302.773438 & 486.498234 & 1112.707788 & 1420.175608 & 1713.995276 & 5096.461763 \\
10 & 7.000000 & 1748.507209 & 1005.698222 & 689.045212 & 1082.160220 & 1333.933966 & 2317.398937 & 3417.452972 \\
20 & 5.000000 & 2351.719235 & 1438.552380 & 170.217090 & 1791.176273 & 2883.562932 & 2967.320745 & 3946.319134 \\
17 & 4.000000 & 1050.753821 & 628.146006 & 144.725591 & 885.747372 & 1252.235905 & 1417.242354 & 1553.817884 \\
14 & 2.000000 & 859.371444 & 643.985999 & 404.004577 & 631.688010 & 859.371444 & 1087.054878 & 1314.738311 \\
19 & 2.000000 & 2694.651342 & 2053.076183 & 1242.907251 & 1968.779296 & 2694.651342 & 3420.523388 & 4146.395433 \\
21 & 1.000000 & 1367.329211 & 0.0 & 1367.329211 & 1367.329211 & 1367.329211 & 1367.329211 & 1367.329211 \\
\bottomrule
\end{tabular}
\caption[Table caption text]{Summary of revenue distribution conditioned on category in our dataset. Sorted by observations count in each category. Notice the means are similar but max values vary by orders of magnitude, which might suggest that some categories are associated with more fat-tailed outcomes than others.}
\label{table:experiments-data-summary-1}
\end{minipage} }
\end{table}

\FloatBarrier

\textbf{Model details}
\begin{itemize}
    \item The model structure we employ is online (minibatch) hierarchical bayesian model. See Appendix \ref{apdx:Model Structure}
    \item We use Horseshoe priors \citep{piironen2017sparsity} for L1-like regularization. This example has low cardinality categoricals and might not need such a powerful regularization, but this regularization technique extends very well to larger feature sets and won’t hurt in smaller feature sets like in this example.
    \item For hierarchical priors loc-scale decentering reparameterization is used \citep{mcelreath2018statistical}.
    \item We use robust scaling for the target to make model optimization and setting priors easier. Straightforward alternative would be standard scaling but we found it hinders the modeling capabilities of such a fat-tailed model, somewhat expectedly.
    \item We implement a truncated Student- distribution by using rejection sampling for the posterior predictive samples. We found it performs better than inverse transform truncation for such a model in terms of computation and in terms of quality of tails density estimation.
    \item For categorical features we use our streaming ordinal encoding with index variable encoding discussed in \nameref{subsec:Streaming Categorical Encoding}. For countinuous features in a model version with larger feature sets (not presented here) we default to online standard scaling.
    \item The training is performed with 3000 train minibatch size, 1500 warmup steps and 500 samples. Extra 500 warmup steps are performed before fitting to each mini-batch after the first. For this example 500 warmup steps is sufficient but for larger feature sets / more drastic concept drifts more steps would be desirable.
    \item We record posterior sample metrics (estimated out-of-sample) with progressive validation.
    \item The implementation is based on NumPyro framework for PPL and bayesian mondeling and River framework for online preprocessing. With our custom extensions. The code for now is available upon request. 
\end{itemize}

\textbf{Results}
\begin{itemize}
\item Student-$t$ model discriminates between tails, showing orders of magnitude difference in maximum values in samples ($10^{4} - 10^{6}$) and degrees of freedom. While Gaussian model expectedly shrinks and generalizes tails to the same $10^{4}$ order of magnitude.
\item The shape of posterior predictive distribution from Student-$t$ model is visually more fitting to the actual data dstribution. 
\item Error metrics such as MSE, LPPD (Log-Pointwise Predictive Density \citep{mcelreath2018statistical}) and distribution location fit are expectedly better for Student-$t$ model, which is another indicator of relevance of fat-tailed model for this problem (or generally - relevance of robust statistic estimators). Notably, MAE is also better for Student-$t$ model.
\item Both models demonstrate signs of convergence in our online learning settings. 
\item Fat-tailedness can be inferred for each observation via estimated degrees of freedom. For brevitiy we demonstrate the estimation generally, for categories.
\end{itemize}
For detailed results see Appendix \ref{apdx:Detailed results}.
\FloatBarrier

\begin{figure}[h]
    \centering
    \includegraphics[width=0.5\linewidth]{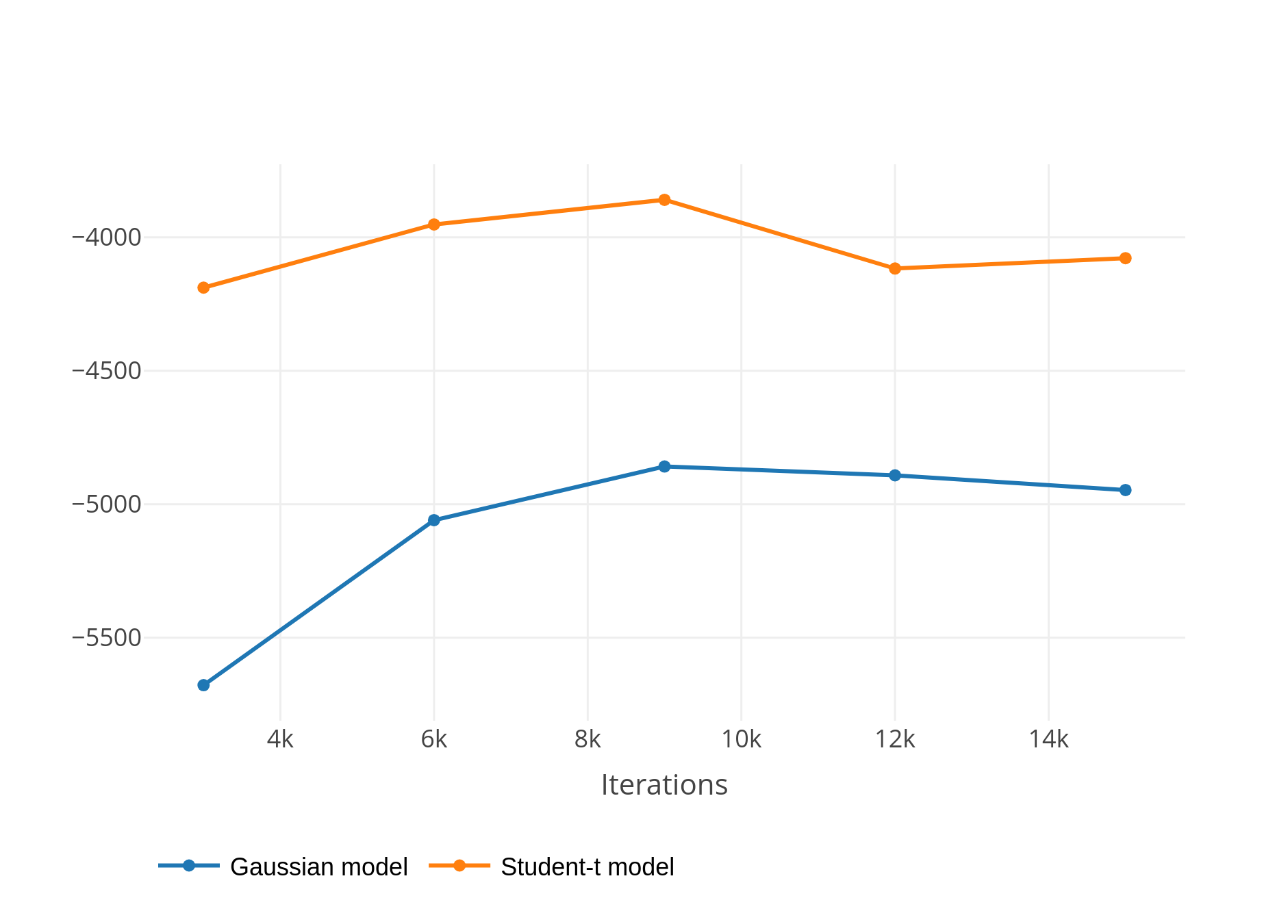}
    \caption{LPPD (on untruncated posterior predictive distribution).}
    \label{fig:log_posterior_predictive_density}
\end{figure}

\begin{figure}[h]
    \centering
    \includegraphics[width=0.5\linewidth]{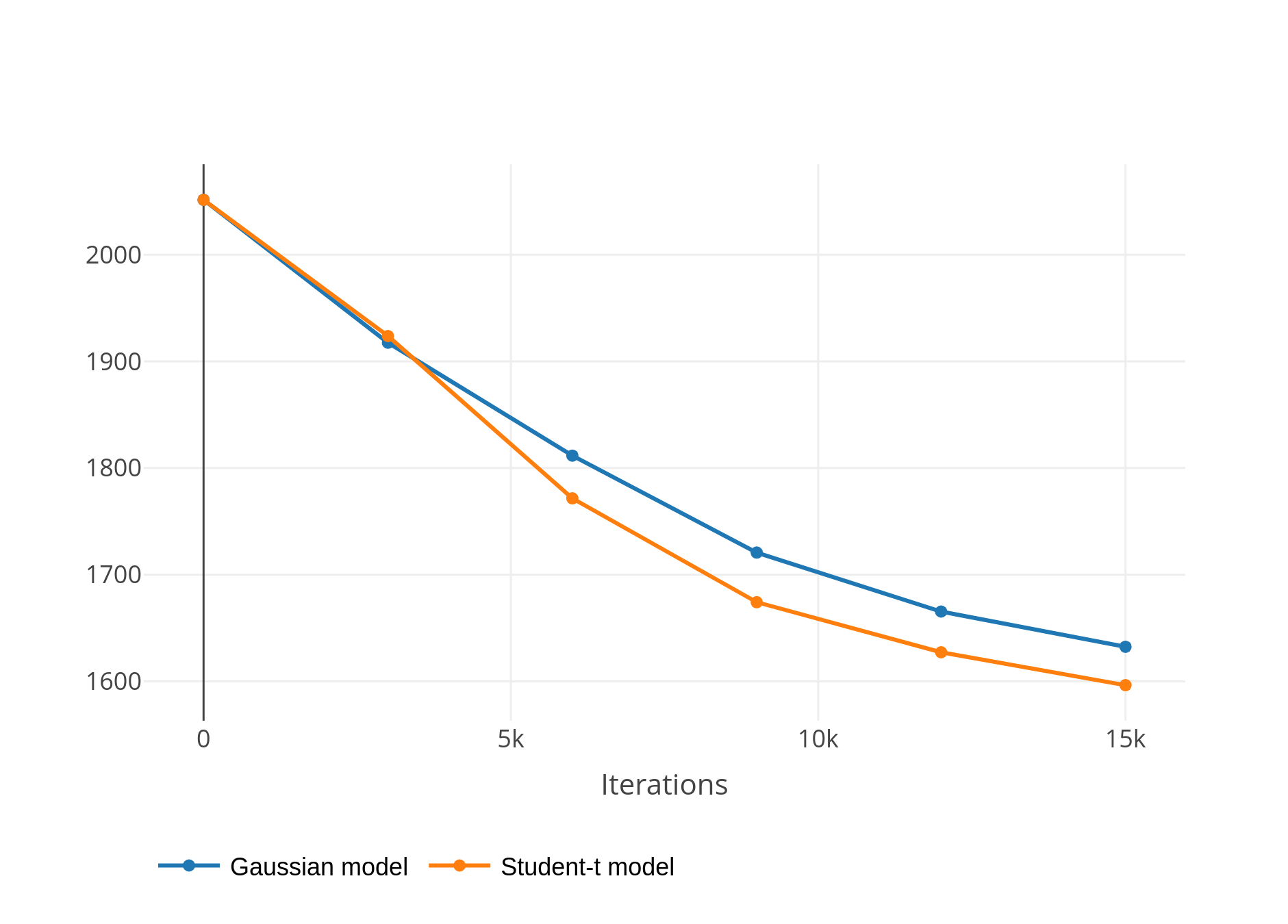}
    \caption{Unscaled RMSE (i.e. with target scaling removed).}
    \label{fig:unscaled root_mean_squared_error loss}
\end{figure}

\begin{figure}[h]
    \centering
    \includegraphics[width=0.5\linewidth]{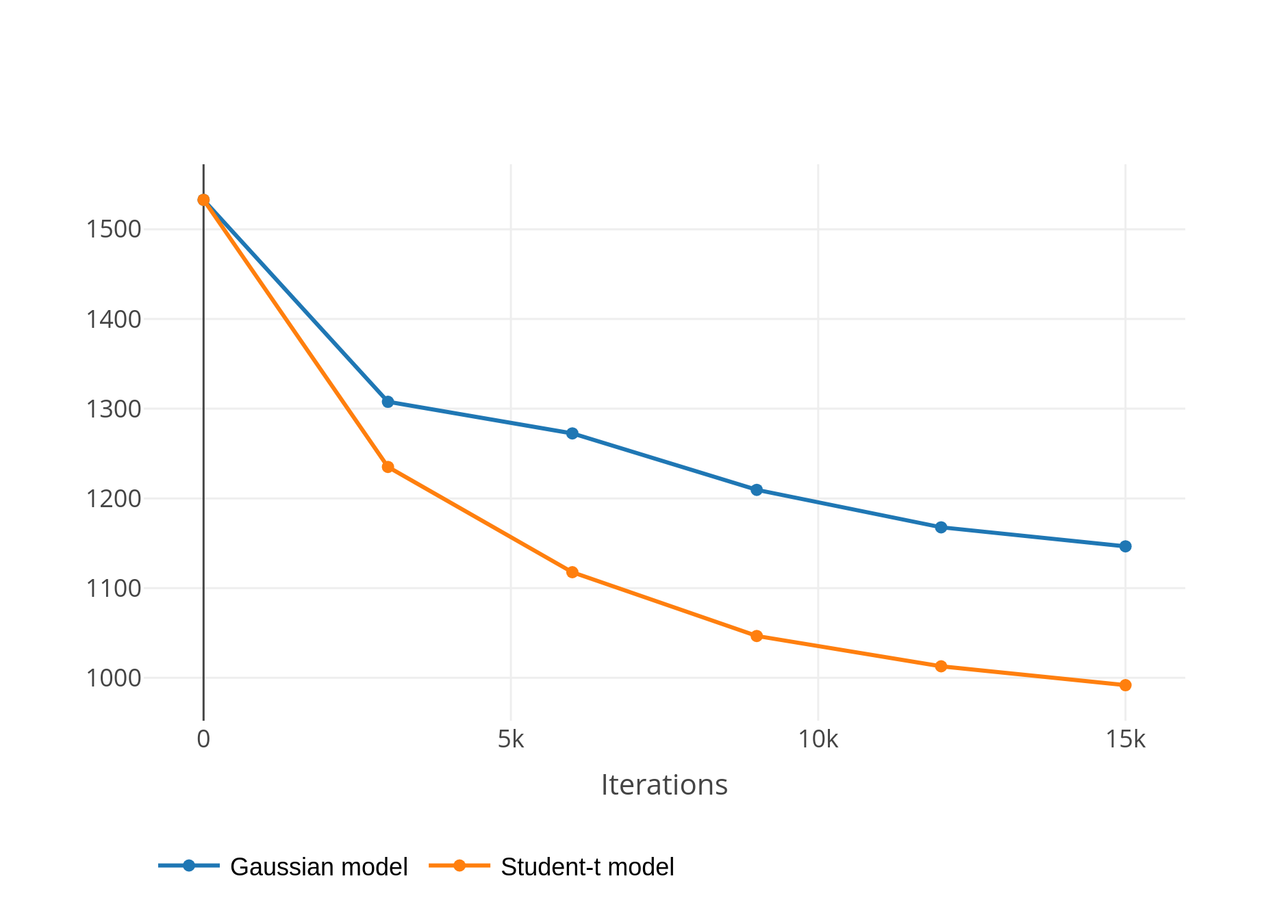}
    \caption{Unscaled MAE (i.e. with target scaling removed).}
    \label{fig:unscaled mean_absolute_error loss}
\end{figure}

\begin{figure}[h]
   \centering
   \begin{tabular}{cc}
       \includegraphics[width=0.5\textwidth]{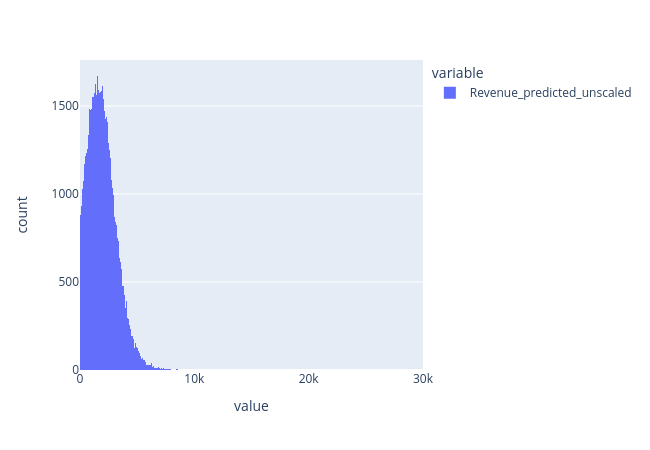} &
       \includegraphics[width=0.5\textwidth]{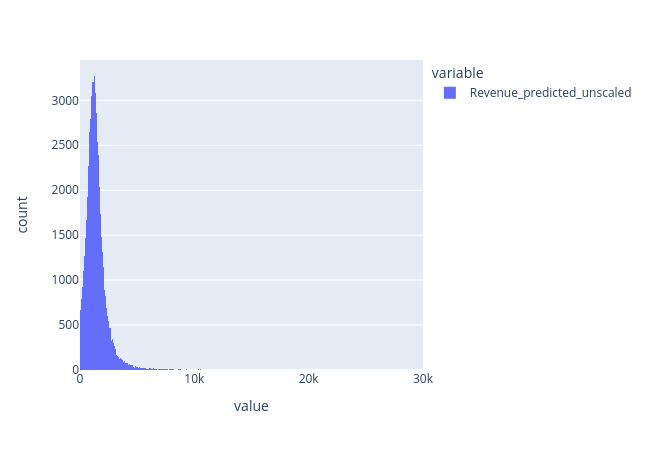}
   \end{tabular}
   \caption{Posterior predictive distribution. Gaussian model (left), Student-$t$ model (right). Compare to the actual data distribution from \ref{fig:experiments-revenue-data-3}}
   \label{fig:Posterior predictive distribution}
\end{figure}

\FloatBarrier

\section{Related work}
\label{sec:Related work}


\textbf{Converting Bayesian Models to Online Models}

For a detailed recent survey on online learning in general with implications of stream learning and adaptive learning with good definitions of terms the reader is refered to \citep{hoi2021online}, for machine perspective in particular - \citep{gomes2019machine}. Good definition of stream learning problem can be found in \citep{gama2013evaluating, domingos2001catching}. For good engineering quality implementations in Python the reader is referred to \citep{montiel2021river}. We focus our discussion on bayesian online learninig and MCMC in particular.

There have been quite some approaches in literature exploring the online learning potential of Bayesian models, particularly incremental learning and scaling to large datasets. Variational Inference (VI) \citep{kucukelbir2015automatic}, Laplace Approximation are computationally powerful alternatives to MCMC in that regard. 

VI can be quite expressive as a building block of complex models like variational autoencoders \citep{kingma2013auto} or Gaussian processess, which can be built with complex combinations of computationally convenient Gaussian distributions. VI provides an analytical solution to an approximation of the posterior. Whereas MCMC provides a numerical approximation to the exact posterior. One of the practical implications of this for modeling approaches we consider (GLMs, hierarchical bayesian models) is that gains in approximation quality (good metrics) from MCMC can be traded for gains in computation speed by switching to VI and vice versa. 

VI based Gaussian processess can be of interest for online learning applications. In fact, there are quite some advancements in that area for adaptive learning and for high-dimensional sparse data \citep{gomez2023adaptive}. But the scalability to high cardinality categorical features for large datasets so fair remains limited and appears to be orders of magnitude more complex than for GLMs and hieararchical bayesian models that do not employ multivariate distributions.

In the current work we focus on MCMC methods due to our attention to exact approximation of non-Gaussian posteriors and robust interpretable models like GLMs and hierachical bayesian models. 

For MCMC in online learning notable ideas include mini-batch algorithm proposed by \citep{neal2011mcmc}, family of Stochastic Gradient MCMC (SG-MCMC) methods, popular in Bayesian Neural Network community, notably SGLD, SGHMC \citep{welling2011bayesian, chen2014stochastic, vadera2020ursabench}. 

We notice in the literature the available approaches having relevant limitations like:

\begin{itemize}
    \item Reliance on decaying step sizes \citep{franzese2021scalable}
    \item Challenges in parametrization of the samplers \citep{franzese2021scalable}
    \item Sample-inefficient exploration of the posterior \citep{neal2011mcmc}
\end{itemize}

SG-MCMC works typically feature multiple passess over the same data. This can provide good convergence under stationary distribution assumptions. But under concept drifts (adaptive learning) - not necessarily. It also conflicts with stream learning settinigs. 

Constraints on decaying/small learning rate are typically undesirable for learning under concept drifts, from our experience. In fact we couldn’t find any literature extensively addressing concept drifts for bayesian MCMC methods (SG-MCMC or any others). Converting a batch MCMC model into a mini-batch in practice proved to be not straightforward under concept drifts. 

Parallel example can be found in Reinforcement Learning literature, where there exist a substantial body of work on multi-armed bandit algorithms with good convergence in stationary conditions. But for non-stationary something with constant learning rate (and worse convergence in stationary) provides better adaptability to concept drifts (and better convergence in non-stationary).

For a survey and formal definitions of concept drifts the reader is referred to \citep{lu2018learning}. The significance of robustness to concept drifts in machine learning practice is also discussed in \citep{hendrycks2021unsolved}.

We used HMC/NUTS sampler as a base for our online sampling algorithm because of it’s efficiency in exploring densities of complex posterior geometries and easy parameterization compared to classic HMC, a quite efficient sampler in itself \citep{neal2011mcmc, hoffman2014no, mcelreath2018statistical}. 

\textbf{Wide data}

Online learning in certain domain like digital advertisement is associated with high cardinality categories, sparsity, wide data. Stream learning from wide data requires either seemingly unbound data structures (example being implementations in River library \citep{montiel2021river}) or projections into a bounded space like hashing trick \citep{agarwal2014reliable}. 

Unfortunately implementing unbounded data structures like in RIver is non-trivial in array/tensor based frameworks which power PPLs (e.g. Numpyro, PyMC). So we just provision extra space in index variable encodings. 

Compared to hashing trick which also uses bounded space and allows for very fast encoding, our encoding scheme maintains direct mapping from codes to categories. This allows us to leverage interpretability and causal relations. 

Bayesian Hierarchical Models in particular have special regularizing opportunities, relevant for sparse data, popular being Horseshoe priors \citep{piironen2017sparsity}. We found the literature on the issue of exploring such methods in stream learning settings to be quite sparse. 

Frequentist online learning methods feature interesting algorithms such as L1-cumulative penalty \citep{tsuruoka2009stochastic} (extended version of L1, especially effective for online learning). We’ve found it to be quite powerful in practice. We implemented an bayesian version of it (both VI and MCMC) and interestingly enough it consistently gave results similar to Horseshoe prior. Horseshoe prior is more straightforward to implement for bayesian models. 

\textbf{Fat Tails}

For an intuition on fat tails the reader is referred to \citep{mandelbrot2010focusing, taleb2007black}. Formal survey can be found in \citep{taleb2020statistical}. Older interesting works include \citep{mandelbrot1997fractals} where several kinds of randomness are defined, particularly Wild Randomness, which corresponds to very fat tails.

The problem of modeling fat tailed variables had recieved attention in LTV literature. In domains of sales forecasting and LTV modeling quantile regression / objectives have been used \citep{makridakis2022m5, benoit2009benefits}. In LTV works exploring the problem of fat tails Lognormal and Gamma distributions have been used, for a recent survey the reader is reeferred to \citep{wang2019deep}. We haven’t found works addressing extremely fat tails like Cauchy or generalizing thin tails and fat tails in LTV domain like we do.



\bibliographystyle{unsrtnat}







\appendix
\section{Model Structure}
\label{apdx:Model Structure}
\begin{figure}[!htbp]
    \centering
    \includegraphics[width=1.0\linewidth]{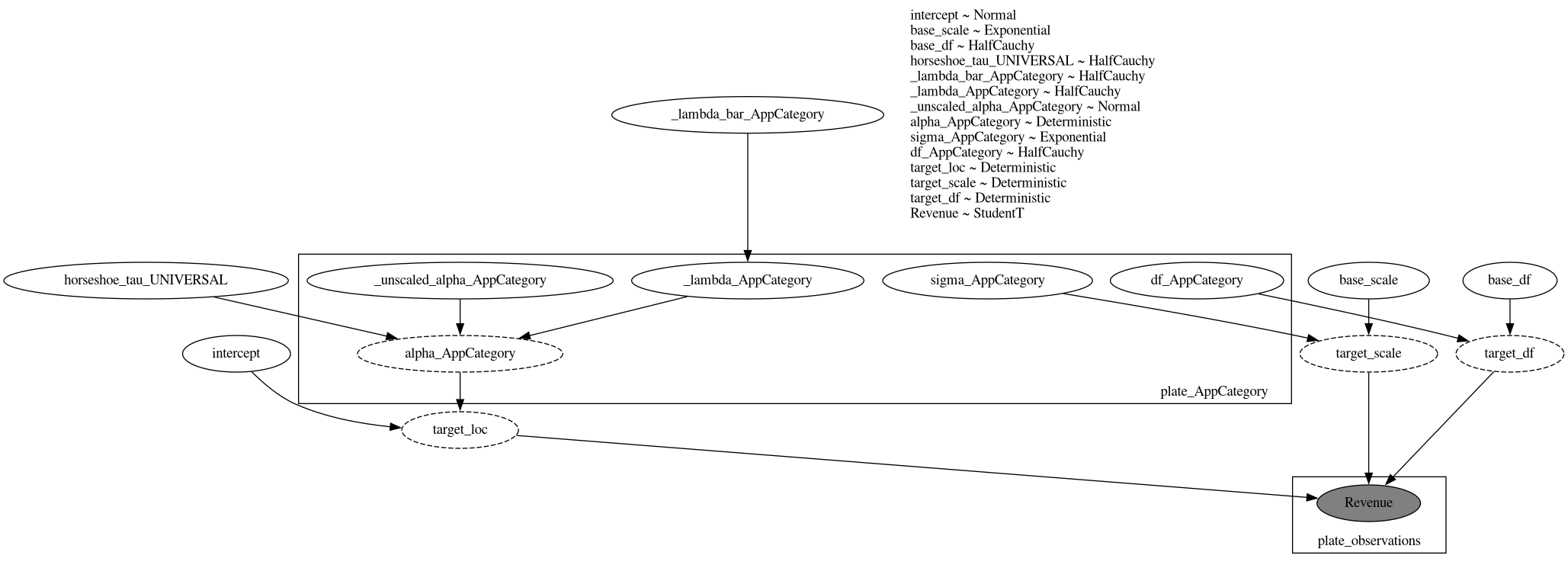}
    \caption{Hierarchical bayesian model with Horseshoe priors for Student-$t$ LTV returns. Degrees of freedom for Student-$t$ were initialized with Half-Cauchy priors to make it easier for Student-$t$ to converge to Gaussian and make it harder to converge to Cauchy, assuming a more conservative estimate.}
    \label{fig:Student-t-model-graph}
\end{figure}

\begin{figure}[!htbp]
    \centering
    \includegraphics[width=1.0\linewidth]{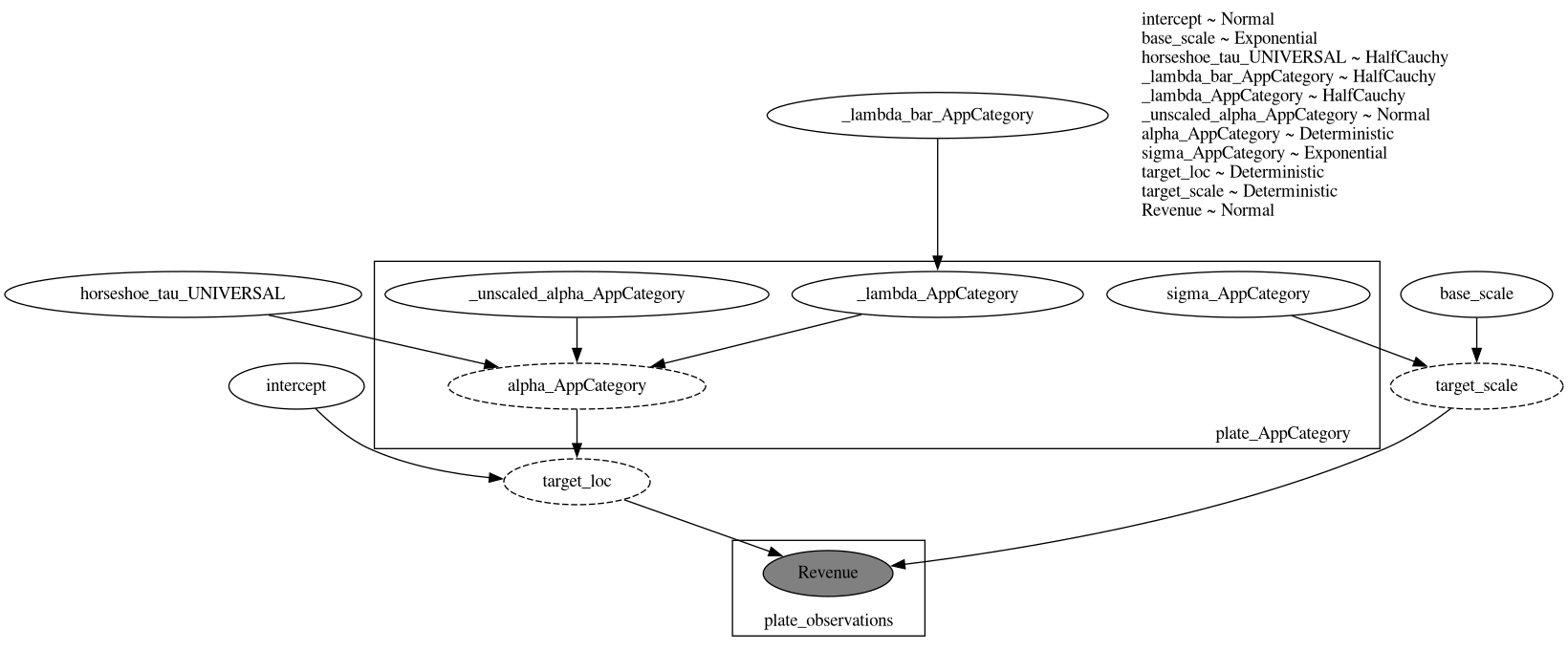}
    \caption{Hierarchical bayesian model with horseshoe priors for Gaussian LTV returns.}
    \label{fig:gaussian-model-graph}
\end{figure}

\FloatBarrier

\section{Detailed results}
\label{apdx:Detailed results}
\setlength{\tabcolsep}{0pt} 
\renewcommand{\arraystretch}{0.7} 
\begin{table}[!htbp]
\resizebox{0.8\textwidth}{!}{\begin{minipage}{\textwidth}
\begin{tabular}{@{}lSSSSSSSS[table-align-exponent = false]@{}}
\toprule
 & {count} & {mean} & {std} & {min} & {25\%} & {50\%} & {75\%} & {max} \\
{AppCategory} & & & & & & & & \\
\midrule
3 & 2598000.000000 & 1484.227086 & 3107.849140 & 0.001628 & 858.927116 & 1255.569689 & 1725.644699 & 2417405.703271 \\
4 & 2060000.000000 & 1476.048143 & 2456.759384 & 0.023249 & 876.451044 & 1269.766774 & 1731.760288 & 2117528.769471 \\
2 & 1411500.000000 & 1422.685571 & 2258.261270 & 0.003755 & 852.495514 & 1247.771379 & 1710.619330 & 1734030.709053 \\
6 & 692000.000000 & 1525.002807 & 2687.521303 & 0.004620 & 902.287246 & 1336.007372 & 1836.661624 & 1266140.569720 \\
9 & 377500.000000 & 1606.947749 & 2919.933486 & 0.002020 & 901.617744 & 1369.214174 & 1923.184481 & 1154495.806402 \\
5 & 92000.000000 & 1384.058395 & 1615.234263 & 0.087212 & 817.479161 & 1220.437023 & 1693.354531 & 324725.628372 \\
8 & 76000.000000 & 1909.612805 & 5569.290965 & 0.103801 & 921.515885 & 1471.766981 & 2182.255655 & 585874.019268 \\
11 & 56000.000000 & 1518.944468 & 2554.392921 & 0.135188 & 860.498375 & 1271.057192 & 1772.594108 & 257869.000737 \\
7 & 51500.000000 & 1731.732735 & 4051.926777 & 0.225049 & 869.792082 & 1337.500749 & 1946.925676 & 405625.765388 \\
15 & 39500.000000 & 1460.997623 & 1661.598603 & 0.080081 & 830.121895 & 1248.367719 & 1753.092450 & 169781.663371 \\
18 & 13500.000000 & 1994.345109 & 4616.065917 & 1.236903 & 939.994848 & 1477.354065 & 2242.468894 & 254692.828121 \\
16 & 9000.000000 & 1746.882078 & 2026.728306 & 0.989973 & 892.980768 & 1398.350765 & 2060.675066 & 55153.211001 \\
13 & 7500.000000 & 1950.845660 & 3374.506018 & 1.972801 & 862.422673 & 1418.979840 & 2190.370961 & 108672.124544 \\
12 & 5500.000000 & 1950.199128 & 2443.204342 & 2.177869 & 933.131484 & 1459.915230 & 2238.258268 & 64295.538698 \\
10 & 3500.000000 & 2093.412779 & 2779.756614 & 3.506331 & 898.537432 & 1468.341640 & 2452.108829 & 87837.480245 \\
20 & 2500.000000 & 2649.690697 & 11249.368951 & 0.418483 & 977.362804 & 1624.961353 & 2854.569979 & 530098.387835 \\
17 & 2000.000000 & 1780.904037 & 1978.437892 & 2.197878 & 879.227302 & 1303.140289 & 1979.921532 & 26930.195424 \\
14 & 1000.000000 & 2188.802852 & 4642.899053 & 5.465640 & 837.337927 & 1439.200658 & 2201.171760 & 87924.812636 \\
19 & 1000.000000 & 2478.844132 & 4168.276719 & 0.740959 & 933.572984 & 1535.587339 & 2528.231561 & 61408.244849 \\
21 & 500.000000 & 2493.620522 & 4299.786561 & 12.014823 & 958.673224 & 1586.522098 & 2609.095279 & 62366.630361 \\
\bottomrule
\end{tabular}
\caption[Table caption text]{Summary of posterior predictive samples distribution per category value. Student-$t$ model. Notice the tails at "max" column.}
\label{table:experiments-student-t-summary-1}
\end{minipage} }
\end{table}

\setlength{\tabcolsep}{0pt} 
\renewcommand{\arraystretch}{0.7} 
\begin{table}[!htbp]
\resizebox{0.8\textwidth}{!}{\begin{minipage}{\textwidth}
\begin{tabular}{@{}lSSSSSSSS[table-align-exponent = false]@{}}
\toprule
 & {count} & {mean} & {std} & {min} & {25\%} & {50\%} & {75\%} & {max} \\
{AppCategory} & & & & & & & & \\
\midrule
3 & 2598000.000000 & 2024.089260 & 1273.127801 & 0.003329 & 1051.377480 & 1865.538228 & 2798.142020 & 11366.847711 \\
4 & 2060000.000000 & 1948.312814 & 1174.525514 & 0.002669 & 1037.312432 & 1825.675664 & 2708.042216 & 8509.180349 \\
2 & 1411500.000000 & 1818.724777 & 1058.551187 & 0.006101 & 1004.496228 & 1722.155586 & 2509.786305 & 8449.620064 \\
6 & 692000.000000 & 1965.008805 & 1226.767405 & 0.004706 & 1043.791658 & 1815.249918 & 2689.956412 & 10787.635570 \\
9 & 377500.000000 & 2208.656591 & 1503.477043 & 0.024526 & 1100.625816 & 1954.970762 & 3003.143655 & 14586.641020 \\
5 & 92000.000000 & 1681.153399 & 952.462454 & 0.108822 & 970.720372 & 1594.279216 & 2276.870949 & 7300.006560 \\
8 & 76000.000000 & 2387.539678 & 1677.622846 & 0.010124 & 1156.099014 & 2099.404617 & 3252.652581 & 14752.420009 \\
11 & 56000.000000 & 1971.828611 & 1329.346695 & 0.092697 & 1072.470460 & 1759.586979 & 2569.286860 & 22745.881677 \\
7 & 51500.000000 & 2166.546192 & 1760.653669 & 0.071778 & 1053.105618 & 1738.153654 & 2700.834264 & 22119.730463 \\
15 & 39500.000000 & 1695.565172 & 1042.745345 & 0.019110 & 951.135501 & 1560.154840 & 2253.272883 & 14955.632660 \\
18 & 13500.000000 & 1926.522852 & 1245.225112 & 0.443840 & 1072.143595 & 1749.459361 & 2557.533089 & 15906.564345 \\
16 & 9000.000000 & 1885.010787 & 1198.048058 & 0.152123 & 1065.541564 & 1718.645020 & 2477.525552 & 13733.428595 \\
13 & 7500.000000 & 2075.616498 & 1503.035284 & 0.080332 & 1106.144313 & 1799.157022 & 2674.653454 & 15161.816893 \\
12 & 5500.000000 & 2053.358609 & 1736.000574 & 1.112706 & 1121.336568 & 1797.645308 & 2627.431234 & 58106.746669 \\
10 & 3500.000000 & 2051.971653 & 1523.758262 & 4.981271 & 1071.875284 & 1795.386487 & 2670.872295 & 24343.344938 \\
20 & 2500.000000 & 2323.015888 & 1686.137605 & 1.149778 & 1197.769318 & 2004.850187 & 3039.309081 & 19488.840586 \\
17 & 2000.000000 & 1993.103457 & 1348.161613 & 4.537261 & 1122.552008 & 1740.579856 & 2485.899289 & 13348.821197 \\
14 & 1000.000000 & 2091.319005 & 1585.173897 & 2.715287 & 1114.000963 & 1789.667679 & 2663.199531 & 16161.293699 \\
19 & 1000.000000 & 2235.242389 & 1849.448063 & 63.506574 & 1109.028962 & 1885.878544 & 2848.203677 & 27522.818232 \\
21 & 500.000000 & 2555.747187 & 2560.732034 & 15.987972 & 1320.601897 & 2187.471090 & 3131.678371 & 32674.953202 \\
\bottomrule
\end{tabular}
\caption[Table caption text]{Summary of posterior predictive samples distribution per category value. Gaussian model.}
\label{table:experiments-gaussian-summary-1}
\end{minipage} }
\end{table}

\setlength{\tabcolsep}{0pt} 
\renewcommand{\arraystretch}{0.7} 
\begin{table}[!htbp]
\resizebox{0.75\textwidth}{!}{\begin{minipage}{\textwidth}
\begin{tabular}{@{}lSSSSSSSS[table-align-exponent = false]@{}}
\toprule
 & {count} & {mean} & {std} & {min} & {25\%} & {50\%} & {75\%} & {max} \\
{$df\_AppCategory$} & & & & & & & & \\
\midrule
$base\_df$ & 1.583213 & 0.224891 & 1.606348 & 1.232769 & 1.947801 & 192.223686 & 1.016369 \\
$df\_AppCategory[0]$ & 3.037992 & 7.394851 & 0.957176 & 0.009343 & 7.231927 & 213.611758 & 1.002431 \\
$df\_AppCategory[1]$ & 18.063458 & 211.786066 & 0.994752 & 0.008960 & 6.377130 & 210.378982 & 1.002759 \\
$df\_AppCategory[2]$ & 0.794834 & 0.371778 & 0.758689 & 0.202816 & 1.374307 & 267.035141 & 1.008266 \\
$df\_AppCategory[3]$ & 0.592255 & 0.280609 & 0.568467 & 0.118086 & 1.013355 & 147.154446 & 1.024842 \\
$df\_AppCategory[4]$ & 0.397004 & 0.254064 & 0.370024 & 0.010627 & 0.732721 & 173.500981 & 1.011555 \\
$df\_AppCategory[5]$ & 1.447047 & 1.493848 & 1.006491 & 0.004400 & 3.173628 & 596.434217 & 0.998016 \\
$df\_AppCategory[6]$ & 0.476433 & 0.295247 & 0.440746 & 0.008905 & 0.878853 & 199.089211 & 1.017914 \\
$df\_AppCategory[7]$ & 0.870743 & 2.542244 & 0.454511 & 0.000732 & 1.749054 & 354.224821 & 1.001101 \\
$df\_AppCategory[8]$ & 0.673858 & 0.690040 & 0.494494 & 0.001701 & 1.393110 & 638.510405 & 1.001232 \\
$df\_AppCategory[9]$ & 0.511213 & 0.373786 & 0.443568 & 0.001489 & 1.019495 & 478.062082 & 1.002069 \\
$df\_AppCategory[10]$ & 6.709053 & 40.865642 & 1.026196 & 0.000778 & 5.545111 & 283.437165 & 0.998912 \\
$df\_AppCategory[11]$ & 3.712551 & 7.334921 & 1.714618 & 0.000146 & 8.857665 & 352.299999 & 0.998524 \\
$df\_AppCategory[12]$ & 45.108482 & 888.939305 & 1.118163 & 0.004221 & 5.680358 & 477.721446 & 0.999921 \\
$df\_AppCategory[13]$ & 10.566993 & 86.615418 & 1.221434 & 0.006384 & 10.870121 & 258.349041 & 1.002204 \\
$df\_AppCategory[14]$ & 3.279646 & 7.802838 & 1.170356 & 0.002383 & 7.156668 & 482.907175 & 1.006677 \\
$df\_AppCategory[15]$ & 0.960589 & 1.796895 & 0.634498 & 0.004607 & 1.798340 & 509.507438 & 0.999906 \\
$df\_AppCategory[16]$ & 5.271386 & 24.173570 & 1.302414 & 0.006118 & 6.061284 & 241.643534 & 1.010686 \\
$df\_AppCategory[17]$ & 6.008043 & 40.374809 & 1.110147 & 0.002681 & 7.541149 & 320.962356 & 1.004294 \\
$df\_AppCategory[18]$ & 16.812221 & 223.456146 & 1.212741 & 0.007181 & 7.236889 & 366.646899 & 1.001436 \\
$df\_AppCategory[19]$ & 3.186932 & 8.790148 & 1.121456 & 0.005001 & 6.079355 & 388.863636 & 0.998084 \\
$df\_AppCategory[20]$ & 3.067101 & 13.033936 & 1.096541 & 0.006277 & 4.837809 & 487.994672 & 1.006234 \\
$df\_AppCategory[21]$ & 2.986581 & 7.913037 & 1.017282 & 0.008583 & 5.623921 & 291.347248 & 1.006925 \\
\bottomrule
\end{tabular}
\caption[Table caption text]{Summary of posterior degrees of freedom of Student-T model for each category. During inference for observation k the degree of freedom is inferred as $base\_df + df\_AppCategory[i]$ where $i$ is the encoded category value for observation k. The percentile intervals are not representative of highest density intervals and should not be used for inference because because priors for $df$ are asymmetric (by design). But some categories indeed look more fat-tailed than others (e.g. where mean $df$ < 1.0)}
\label{table:Summary of posterior degrees}
\end{minipage} }
\end{table}

\begin{figure}[!htbp]
   \centering
   \begin{tabular}{cc}
       \includegraphics[width=0.5\linewidth]{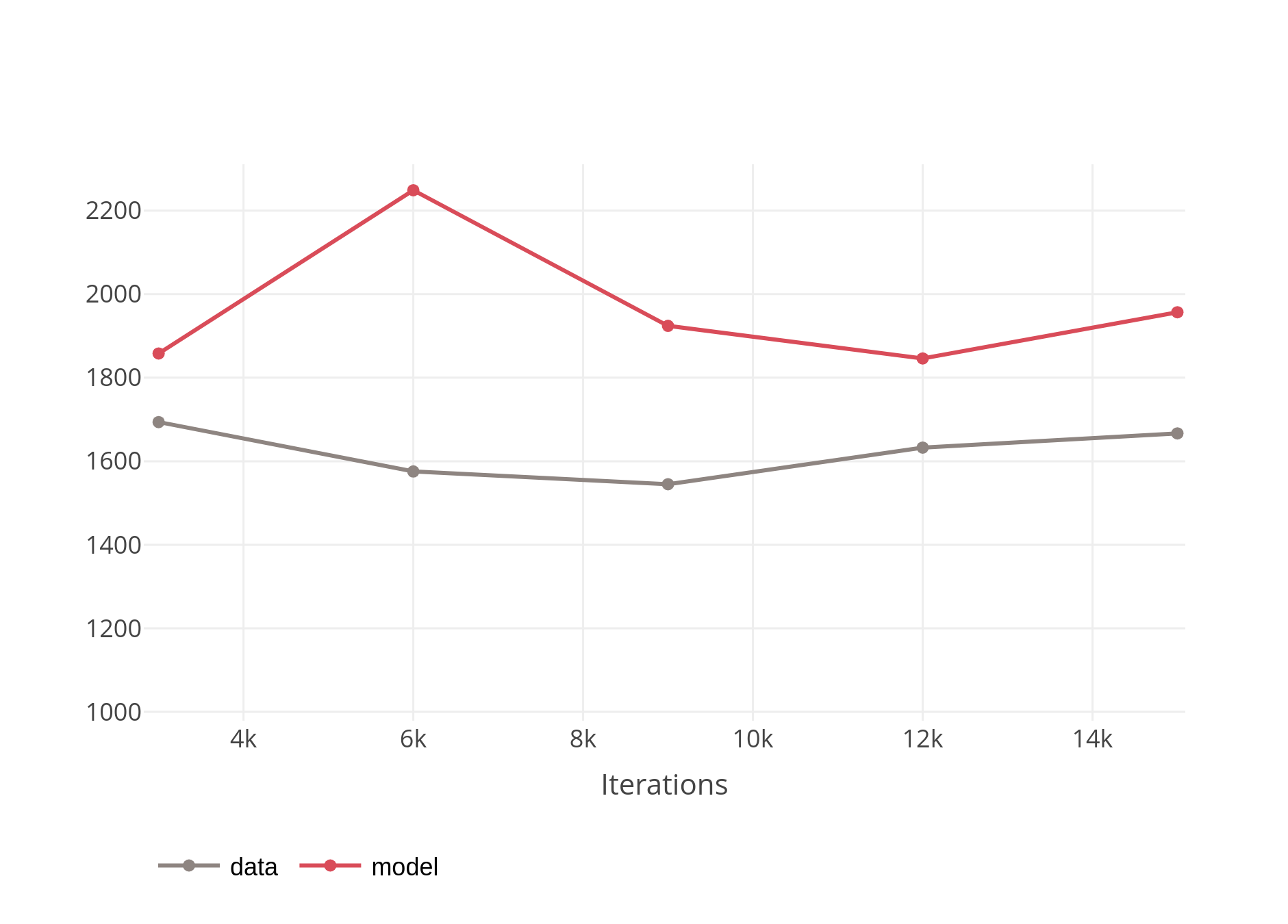} &
       \includegraphics[width=0.5\linewidth]{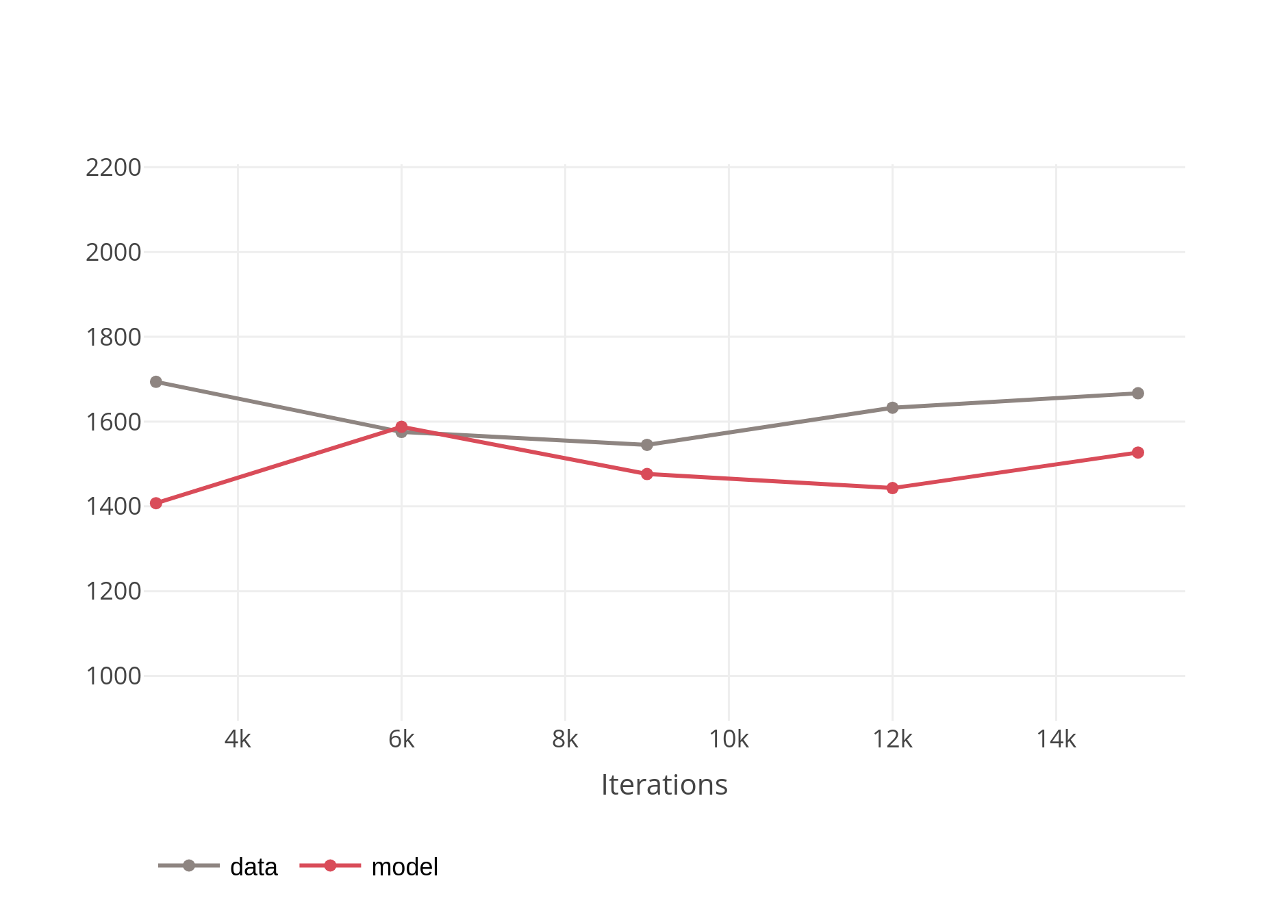}
   \end{tabular}
   \caption{Distribution location fit. Gaussian model (left), Student-$t$ model (right). We use this plot to check adaptability to concept drifts in posterior distribution location. In this example drifts are not drastic for drift adaptability to be evident. But notice that due to fat-tailed nature in the data the location estimate for Gaussian model is tilted to higher values (more biased than Student-$t$).}
   \label{fig:Distribution location fit}
\end{figure}

\end{document}